\def\year{2019}\relax
\documentclass[letterpaper]{article} 
\usepackage{aaai19}  
\usepackage{times}  
\usepackage{helvet}  
\usepackage{courier}  
\usepackage{url}  
\usepackage{graphicx}  
\usepackage{xcolor}
\usepackage{color}

\usepackage{enumerate}
\usepackage{float}
\usepackage{hyperref}       
\usepackage{booktabs}       
\usepackage{amsfonts}       
\usepackage{nicefrac}       
\usepackage{microtype}      
\usepackage{soul}
\usepackage{subfigure}
\usepackage{multirow}
\usepackage{amsmath}
\usepackage[ruled]{algorithm2e}
\definecolor{khaki}{RGB}{236,165,105}
\definecolor{pinkMy}{RGB}{255,51,204}
\definecolor{brilliantBlue}{RGB}{0,176,240}

\begin{document}
%
\title{FineFool: Fine Object Contour Attack via Attention}
\author{Jinyin Chen*, Haibin Zheng, Hui Xiong, Mengmeng Su\\
College of Information Engineering, Zhejiang University of Technology\\
288 Liuhe Road, Hangzhou, Zhejiang Province, 310023\\
*Corresponding author 
}
\maketitle
\begin{abstract}
Machine learning models have been shown vulnerable to adversarial attacks launched by adversarial examples which are carefully crafted by attacker to defeat classifiers. Deep learning models cannot escape the attack either. Most of adversarial attack methods are focused on success rate or perturbations size, while we are more interested in perturbation's relationship with the image itself. In this paper, we put forward a novel adversarial attack based on contour, named FineFool. Finefool not only has better attack performance compared with other state-of-art white-box attacks with higher attack success rate and smaller perturbation, but also is capable of visualization the optimal adversarial perturbation via attention on object contour. To the best of our knowledge, Finefool is for the first time combines the critical feature of the original clean image with the optimal perturbations in a visible manner. Inspired by the correlations between adversarial perturbations and object contour, slighter perturbations is produced via focusing features on object contour, which is more imperceptible and hard to be defended, especially network add-on defense methods with the trade-off between perturbations filtering and contour feature loss. Compared with existing state-of-art attacks, extensive experiments are conducted to show that Finefool is capable of efficient attack against defensive deep models.
\end{abstract}

\section{Introduction\label{introduction}}
In recent years, deep learning has shown outstanding performance in many professional tasks, including computer vision, bioinformatics, complex networks, natural language processing, etc. However, the shortcomings of deep learning are gradually exposed with its wide application. One of the serious drawbacks in deep learning is the vulnerability of deep models against adversarial examples. For instance, adversarial image, produced by adding carefully designed perturbations to clean image, will be misclassified by state-of-the-art deep model in image classification task. To make matters worse, the perturbations of effective adversarial example are very slight and invisible to human visual system.

With the study extends, adversarial attacks against deep models are gradually systematized, e.g. white/black/semi-black box attacks based on the exposure degree of deep model, or targeted/untargeted attacks based on requirements of the adversary. Moreover, seminal works on adversarial examples fooling deep models exists not only in digital space but also in physical world~\cite{evtimov2017robust}.

The \emph{white-box attack} assumes a full exposure of target deep model, including its parameters, model structure, training process, and even training data. The \emph{black-box attack} denotes adversarial examples (during testing) generated by model output without the knowledge of that model. And \emph{semi-black-box attack} is between white-box and black-box attacks, assuming that adversary has limited knowledge of deep model (e.g. training process and/or model structure), but certainly does not know its parameters. Generally speaking, the perturbations of white-box attack are carefully designed by gradient back propagation, which is slight but weak transferability against different deep models. And adversarial examples of black-box attack are usually produced via the optimization methods, which have stronger attack transferability but slightly larger perturbations.

In order to illustrate targeted/untargeted attacks clearly, given a classifier $f(\Theta, x)$: $x\in\mathcal{X}\to{y}\in\mathcal{Y}$, and clean input $x\in\mathcal{X}$ at first, where $\Theta$ denotes deep model parameters. The task of \emph{untargeted attack} can be expressed as: $f(\Theta, x)=y$, $f(\Theta, x^{*})\neq{y}$, where $y$ is ground-truth label of clean example $x$, and $x^{*}$ is adversarial example crafted by adding slight perturbations to $x$. Similarly, adversarial examples generation for \emph{targeted attack} against deep model satisfies the condition: $f(\Theta, x)=y$, $f(\Theta, x^{*})=y'$, where $y'$ is specific target label preset by adversary, and $y'\neq{y}$. Moreover, the $L_{p}$ norm of perturbations is not allowed to be greater than preset threshold $\epsilon$, which can be denoted as $||x^{*}-x||_{p}<\epsilon$.

As mentioned above, adversarial attacks cause misclassification of deep models by adding perturbations to clean examples. And adversarial attacks can severely destroy the performance of deep models, which threaten the security of deep learning-based applications and limit its further research. There are various opinions in the literature about vulnerability of deep models to carefully designed adversarial perturbations. \cite{goodfellow2014explaining} first proposed the view-point that the design of modern
deep models that forces them to behave linearly in high dimensional spaces also makes them vulnerable to adversarial attacks. Furthermore, \cite{tanay2016boundary} proposed the `boundary tilting' perspective, which refuted the linear hypothesis. \cite{cubuk2017intriguing} consider that the ``origin of adversarial examples is primarily due to an inherent uncertainty that neural networks have about their predictions''. \cite{rozsa2016towards} argue that vulnerability of deep model is caused by the training stagnation of its decision boundary. And flatness of decision boundaries, large local curvature of decision boundaries and low flexibility of deep models are other interpretations for adversarial attacks against deep models~\cite{akhtar2018threat}.

Since deep model has outstanding capacity of critical feature extracting, the extracted feature mainly locates on object contour. We believe there are hidden relationships between carefully designed adversarial perturbations and original clean image, especially the object contour in the image. Moreover, what is the correlation between perturbations and feature maps in hidden layers of deep model? Motivated by the question, we visualize the extracted object contour, pay attention to the critical local feature, update perturbations via back propagation gradient and forward propagation attention map. We propose a novel adversarial attack method FineFool with slighter perturbations faster on the premise of guaranteeing attack effect. Besides, Finefool is an efficient tool to evaluate the defensibility of deep model, and can improve its defensibility by adversarial training~\cite{miyato2016adversarial,miyato2017virtual,madry2017towards}. What's more, attention mechanism of feature maps in deep model is used to figure out attention map, which consists of channel spatial attention and pixel spatial attention. And visualization is applied to observe relativity of feature maps and attention map with perturbations, and interpret deep models~\cite{koh2017understanding,dong2017towards}.

As opposed to attack, adversarial defense is also a research focus. There are mainly three types of defense methods, including input modification, network structure modification and network add-on. The implementation of input modification defense method destroys carefully designed perturbations by modifying input data, e.g. discrete re-encoding, adding random noise, or image flip/rotation. And network structure modification is applied to improve network robustness by modifying network architecture, such as activation function, polling layer, convolution layer, etc. The network add-on defense method keeps the original deep model intact and appends external model to it during testing, e.g. perturbation rectifying network~\cite{akhtar2017defense}. But FineFool can effectively beat deep model with defensive network add-on, which defenses against adversarial examples by filtering out perturbations. If keeping up object contour with classification feature, the perturbations crafted by attention focus of object contour can not be completely filtered. Conversely, filtering perturbations generated by FineFool will cause a great loss of object contour with classification information. Hence, the trade-off between perturbations filtering and object contour keeping determines attack ability of FineFool against network add-on defense.

In summary, main contributions of this paper can be summarized as follows:
\begin{itemize}
\item Compared with the current state-of-the-art attacks, the perturbations of FineFool can be further reduced while maintaining similar attack success rate. And slighter perturbations is implemented by focusing features onto object contour, which is the first time introducing attention mechanism into adversarial attacks to the best of our knowledge.
\item An important observation has been made by visualization that there are strong correlations between perturbations and object contour, feature maps, attention map. And perturbations crafted by FineFool is an inseparable part of original image, which is more invisible. The observation provides a new research direction for adversarial examples generation.
\item Intensive experiments on ImageNet dataset show that the proposed method is attack effective against different deep models with defensive network add-on, which can be applied to evaluate robustness of deep models and train a robust model via adversarial training.
\end{itemize}

The rest of the paper is organized as follows. The related works are discussed in Section 2. The FineFool method and critical techniques are introduced in Section 3. The experiments and conclusions are presented in Section 4 and 5, respectively.

\section{Related Works\label{relatedworks}}
Here we summarize related works on attacking and defending deep models against adversarial examples, and attention mechanism.

\subsection{Attacks to DNN}
There are mainly two types of attacks based on transparency degree of DNN model, gradient-based and optimization-based attacks, corresponding to white-box and black-box attacks,respectively. The white-box attack methods applied in experiments is described in detail below.

\subsubsection{Fast Gradient Sign Model (FGSM):}~\cite{goodfellow2014explaining} proposed the simplest and most widely used untargeted attack, which applied backward propagation gradient from the target DNN to generate adversarial examples. Perturbation is evaluated by $\rho =\epsilon sign(\bigtriangledown J(\theta,I_{c},l))$, where $\bigtriangledown{J}$ denotes the gradient of the original image $I_{c}$ around the model parameters $\theta$. $sign(.)$ denotes the sign function, and $\epsilon$ is a small scalar value that limits the disturbance norm.

\subsubsection{MI-FGSM:}~\cite{dongboosting} introduced a broad class of momentum-based iterative algorithms to boost adversarial attack ability, which can stabilize update directions and escape from poor local maxima during the iterations via integrating the momentum term into the iterative process for attacks. And the momentum iterative algorithm was applied to an ensemble of deep models, which show that the adversarially trained models with a strong defense ability are also vulnerable to MI-FGSM.

\subsubsection{Projected Gradient Descent (PGD):}~\cite{carlini2017towards} mentioned the standard method for large-scale constrained optimization, PGD, which performs one step of standard gradient descent, and then clips all the coordinates to be within the box. \cite{madry2017towards} demonstrated the local maxima found by PGD all have similar loss values, both for normally trained networks and adversarially trained networks. This concentration phenomenon suggests that robustness against the PGD adversary yields generalization of robustness.

\subsubsection{Basic Iterative Model (BIM):}~\cite{kurakin2016adversarialPhysical} demonstrated a standard convex optimization method, BIM, which is equivalent to projected gradient descent. It is an extension of the single-step method, which takes multiple small step iterations while adjusting the direction after each step. After a sufficient number of iterations, BIM can successfully generate an adversarial example classified into the target label.

\subsubsection{DeepFool:}~\cite{moosavi2016deepfool} proposed DeepFool, which is a simple but very effective untargeted attack. In each iteration, it calculates the minimum distance $d(y_{1},y_{0})$ required for each label $y_{1}\ne y_{0}$ to reach the class boundary by approximating the model label with a linear label. $y_{1}$ represents the label of the adversarial example classified by deep model, and $y_0$ represents the true label of the adversarial example. Then make the appropriate steps in the direction of the nearest class. The image perturbation for each iteration are accumulated and the final perturbation is calculated once the output criteria changes.

\subsubsection{Carlini and Wagner Attacks (C\&W):}~\cite{carlini2017towards} designed C\&W attack, which is one of the strongest attacks. Its attack essence is a kind of refined iterative gradient attack that uses the Adam optimizer. It uses the internal configuration of the target DNN to guide the attack, and uses $L_{2}$ specification to quantify the difference between the hostile and original image.

\subsection{Defenses in DNN}
With the continuous expansion of deep learning, its vulnerability to adversarial attacks needs to be addressed urgently. Currently, there are several essential explanations for adversarial examples, including linearity hypothesis, high-nonlinearity, subspace intersection, boundary tilting, inherent prediction uncertainty, evolutionary stalling and so on~\cite{akhtar2018threat}. And different methods to defend against attacks have been put forward based on the different understandings of the essence.

According to implementation, there are three types of defense methods including modified training/input, modifying network and network add-on.

\subsubsection{Modified Training/Input:}~\cite{szegedy2013intriguing} first proposed adversarial training with mixed data to improve defense against specific attack. \cite{miyato2016adversarial} and \cite{zheng2016improving} designed virtual adversarial training and stability training to enhance defense effect, respectively. There are many modification methods of training data, including down-sampling~\cite{guo2017countering}, high-frequency filtering~\cite{das2017keeping}, data compression~\cite{Arjun2017Enhancing}, random scale transformation~\cite{xie2017adversarial}, thermometer encoding~\cite{Buckman2018Thermometer} etc.

\subsubsection{Modifying Network:}~Network modification refers to improving defense by modifying structure or parameters. Such methods include Deep Contractive Networks (DCNs)~\cite{gu2014towards}, gradient regularization~\cite{ross2017improving}, defensive distillation~\cite{papernot2016distillation}, DeepCloak based on masking layer~\cite{gao2017deepcloak}, HyperNetworks with statistical filtering~\cite{sun2017hypernetworks} etc. \cite{nayebi2017biologically} studied the relationship between vulnerability of DNNs and the linearity of activation. \cite{madry2017towards} provided a novel perspective based on robust optimization. And \cite{Wan2018Rethinking} improved defense performance by designing a new loss function.

\subsubsection{Network Add-on:}~The defense is implemented by connecting one or more external expansion models in maintaining integrity of model structure. The common expansion models include Perturbation Rectifying Network (PRN)~\cite{akhtar2017defense}, external multi-detectors~\cite{meng2017magnet} and GAN framework~\cite{lee2017generative}.

\subsection{Attention Mechanism for DNN}
Recently, attention mechanisms achieve state-of-the-art performance in deep learning that capture global dependencies, especially tackling sequential decision tasks, e.g. spatio-temporal attention model with long short term memory is applied to human action recognition. \cite{xu2015show} demonstrated the first visual attention model in video captioning, which can pay close attention to critical information of object. \cite{yang2016stacked} applied a stacked spatial attention model to text generation and public opinion prediction, which produced the second attention layer via modulating the attention map of the first one.

Different from sequential decision tasks, attention mechanism also plays a critical role in the region of computer vision with image processing tasks. \cite{wang2017residual} introduced attention mechanism into convolutional neural network, residual attention network, can improve the performance based on trained Residual Network in an end-to-end training fashion. \cite{zhang2018self} proposed the self-attention generative adversarial network to generate more refined images, which traversed feature map with attention-driven to focus on spatial critical information. \cite{chen2017sca} introduced a novel convolutional neural network, SCA-CNN, which incorporated spatial and channel-wise attentions in convolutional neural network. And inspired by this, critical information concerns based on attention mechanism are applied to optimize the perturbation distribution in this paper.

\begin{figure*}[t]
\centering
\includegraphics[width=\linewidth]{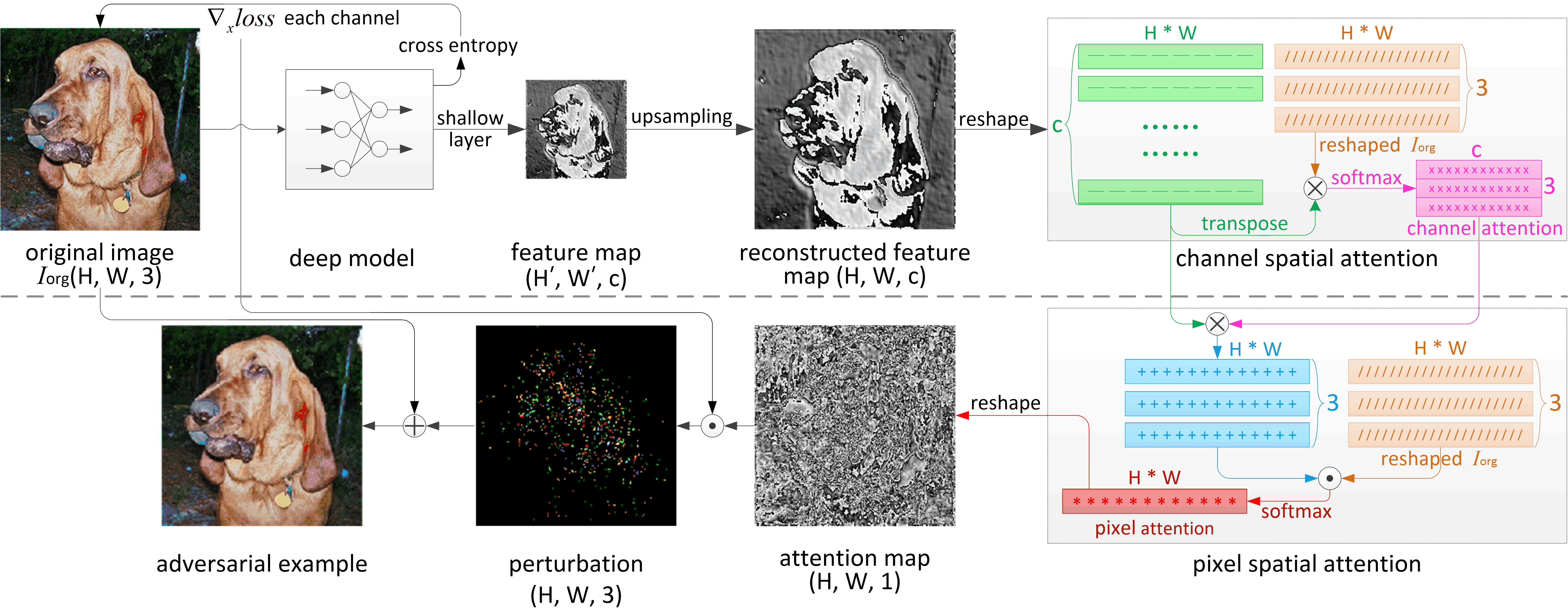}
\caption{Framework of the FineFool attack method. `Deep model' denotes model to be attacked, `feature map' denotes output of deep model in shallow feature layer, `perturbation' is magnified 255 times for better visualization. The channel spatial attention function $\Phi_{c}$ is used to obtain the channel spatial attention weight $W_{c}$, which are multiplied in channel of the feature map. Then, pixel spatial attention function $\Phi_{p}$ is used to obtain the pixel spatial attention weights $W_{p}$, which are multiplied in each pixel points, resulting in an attention map.}
\label{fig:framework_finefool}
\end{figure*}

\section{FineFool Method\label{method}}
The effectiveness of adversarial examples consists of attack ability and perturbations, and strength of attack ability depends on perturbation size and perturbation distribution. Generally speaking, the greater the perturbations, the stronger the attack ability. However, the characteristics of effective adversarial examples are smaller perturbations and stronger attack ability. Therefore, the task of generating effective adversarial examples is a contradictory multi-objective optimization problem. Two issues are studied in this paper, how to search for appropriate position to add perturbations, and how to calculate imperceptible perturbations. Driven by this research mission, a novel approach to adversarial examples generation, FineFool, is proposed, which can improve the size and distribution of perturbations. The framework of FineFool is shown in Figure~\ref{fig:framework_finefool}, where deep model will be attacked, feature map is output of the deep model in shallow feature layer, bilinear interpolation is applied to upsampling, and perturbations are magnified 255 times for better visualization. Under the premise of guaranteeing attack ability, iterative optimization is used to calculate smaller perturbation size, and attention mechanism is used to search more suitable perturbation distribution.

\subsection{Perturbation Distribution}
\begin{figure}[h]
\centering
    \subfigure[original]{
        \includegraphics[width=0.28\linewidth]{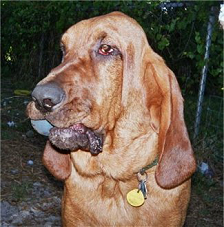}  }
    \subfigure[FineFool]{
        \includegraphics[width=0.28\linewidth]{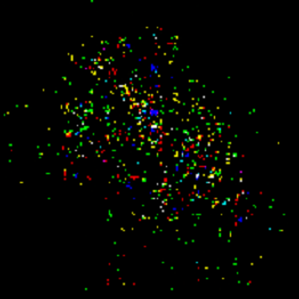}  }
    \subfigure[DeepFool]{
        \includegraphics[width=0.28\linewidth]{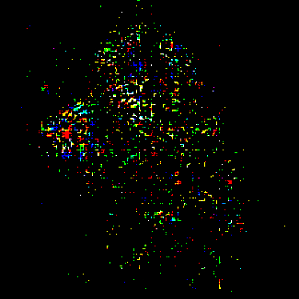}  } \\
    \subfigure[MI-FGSM]{
        \includegraphics[width=0.28\linewidth]{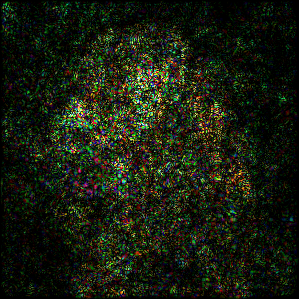}  }
    \subfigure[BIM]{
        \includegraphics[width=0.28\linewidth]{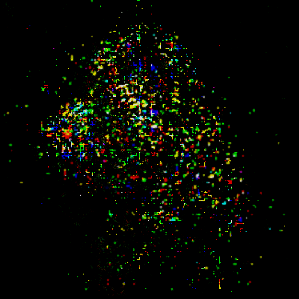}  }
    \subfigure[C\&W]{
        \includegraphics[width=0.28\linewidth]{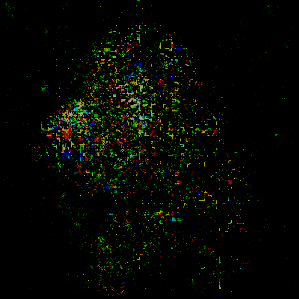}  }
\caption{The visual relationship between original clean image and adversarial perturbations obtained by different gradient-based attacks. And both perturbations are magnified 255 times for better visualization. Compared to other attack methods, the perturbations obtained by FineFool is minimal, which shows better concealment.}
\label{fig:perturbationVisualization}
\end{figure}

The existence of adversarial examples is currently considered to be due to local linearity of deep model, resulting the model to be sensitive to changes in input. Firstly, one observation in massive experiments is analyzed that gradient-based adversarial perturbation is mainly concentrated near the target object, especially the contour of the object, as shown in Figure~\ref{fig:perturbationVisualization}. This is not difficult to understand that convolution kernel is essentially a filter and different convolution kernels extract different features of the same image. Shallow convolution kernels extract basic texture features, and deep layers extract abstract features. As depth of the model increases, the receptive field of the feature layer continues to increase.

The gradient can be regarded as a quantitative assessment of the sensitivity of input pixel to output confidence. The background is filtered out in forward propagation due to the lack of effective texture features. Contrarily, the target object is retained during forward propagation due to the rich texture features surrounding its contour, which avoid gradient dispersion in backward propagation. That is why gradient-based adversarial perturbations are mostly concentrated on the target object.

To prevent the added perturbations from being filtered by the defense method (e.g. network add-on), FineFool attempts to focus the perturbations on classification feature distribution with contour information. How to capture the outline features of an object? The attention mechanism associated with the feature map layer of neural network $f$ is a good choice.

\subsection{Spatial Attention}
Attention mechanisms are mainly divided into two types: soft attention and hard attention. Hard attention is a stochastic process, which assigns weights based on Bernoulli distribution. In contrast, the soft attention mechanism is a parametric weighting method that can be embedded into neural network structure for end-to-end training, which has achieved good performance in deep models that required global information.

Compared with the shallow features, the deep features have a larger visual field, but spatial information of deep feature maps is greatly lost. Therefore, the shallow feature output of the neural network $f$ is reconstructed to the same dimensional space as the input via bilinear interpolation. The different channel outputs of feature map correspond to the filtering results of different convolution kernels. The attention mechanism for perturbation distribution searching consist of channel spatial attention and pixel spatial attention, where channel spatial attention focuses on feature distribution by weighting the different channels of feature map, and pixel spatial attention focuses on pixel region by weighting the different pixels of feature map. The specific calculation process is as follows.

\subsubsection{Channel Spatial Attention:}
The channel spatial attention weight is represented as pink rectangle in Figure~\ref{fig:framework_finefool}, which is defined as:
\begin{equation}
W_{c}=softmax(I_{org}^{re}\otimes{fm^{T}})
\label{equ:Eq_channelSpatialAttention}
\end{equation}
where $I_{org}^{re}\in\mathbb{R}^{3\times{l}}$ is reshaped original image $I_{org}\in\mathbb{R}^{H\times{W}\times{3}}$,which is shown in channel spatial attention block of Figure~\ref{fig:framework_finefool} with khaki. And $fm\in\mathbb{R}^{c\times{l}}$ denotes reshaped reconstructed feature map, which is shown in channel spatial attention block with green. And $fm^{T}\in\mathbb{R}^{l\times{c}}$ represents transposed $fm$, $l=H\times{W}$, $H$ and $W$ are the height and width of original image, $c$ denotes the channel depth of feature map layer, $\otimes$ represents matrix multiplication. After completing the matrix multiplication of $I_{org}^{re}$ and $fm^{T}$, output channel spatial attention weight $W_{c}\in\mathbb{R}^{3\times{c}}$, which is shown in channel spatial attention block with pink.


\subsubsection{Pixel Spatial Attention:}
The pixel spatial attention weight is represented as red rectangle in Figure~\ref{fig:framework_finefool}, which is defined as:
\begin{equation}
W_{p}=softmax(W_{c}^{re}\odot{I_{org}^{re}})
\label{equ:Eq_pixelSpatialAttention}
\end{equation}
where $W_{c}^{re}=W_{c}\otimes{fm}$ and $W_{c}^{re}\in\mathbb{R}^{3\times{l}}$, which represents reconstructed channel spatial attention weight with blue in pixel spatial attention block. $\odot$ means that matrix $W_{c}^{re}$ and $I_{org}^{re}$ are multiplied by corresponding elements and then summed by columns. And $W_{p}\in\mathbb{R}^{1\times{l}}$ is the output of pixel spatial attention block, which is red in Figure~\ref{fig:framework_finefool}.

Then the pixel attention weight $W_{p}\in\mathbb{R}^{1\times{l}}$ is reshaped to attention map $W_{map}\in\mathbb{R}^{H\times{W}\times{1}}$, which is applied to focus the perturbations on the critical classification features of original image. And the perturbation $\rho\in\mathbb{R}^{H\times{W}\times{3}}$ is defined as:
\begin{equation}
\rho=W_{map}\odot\frac{\nabla_{x}{J(I_{org}, y)}}{||\nabla_{x}{J(I_{org}, y)}||_{1}}
\label{equ:Eq_perturbations}
\end{equation}
where $i=1,2,3$, which corresponds to the RGB channel of original image. And the adversarial examples $x^{*}$ is defined as:
\begin{equation}
x^{*}=I_{org}\oplus{\rho}
\label{equ:Eq_adversarialExamples}
\end{equation}

Furthermore, iterative optimization is applied to calculate fine perturbations, and momentum-based method is used to improve attack ability against deep model. The more details of FineFool algorithm with momentum are presented in Algorithm 1. The calculation of attention map $W_{map}$ takes advantage of local information in feature map, which can be combined with black-box or white-box attacks to produce adversarial examples. And the loss function $J$ is defined as:
\begin{equation}
\begin{array}{ll}
J(x,y)= & max(-\kappa, Z(x)_{y}-max\{Z(x)_{y^{'}}:y^{'}\neq{y}\} ) \\
    & +||x-x_{0}||_{2}^{2}
\end{array}
\label{equ:Eq_lossJ}
\end{equation}
where $\kappa\geq{0}$ is the hyper-parameter. The larger the value of $\kappa$, the higher the reliability requirement for adversarial examples. $x_{0}$ denotes original clean image, $Z$ is the confidence output of $f$, $||x-x_{0}||_{2}^{2}$ is a regular term used to constrain the perturbation size.

\begin{algorithm}[h]
\caption{FineFool.}
\LinesNumbered
\KwIn{A classifier $f$ with loss function $J$; the iteration step length $\alpha$; the number of iterations $T$; the size of perturbation $\epsilon$; original clean example $x$ and corresponding label $y$; and decay factor $\mu$.}
\KwOut{Adversarial example $x^{*}$ with $||x^{*}-x||_{\infty}\leq{\epsilon}$.}
$g_{0}=0; x_{0}^{*}=x$\;
\While{$i < T$}{
    Input $x_{i}^{*}$ to $f$ and obtain the gradient $\nabla{J(x_{i}^{*}, y)}$ and feature map\;
    Obtain $fm_{i}$ by upsampling feature map\;
    Calculate $W_{c}(i)$ based on Eq.~(\ref{equ:Eq_channelSpatialAttention})\;
    Calculate $W_{p}(i)$ based on Eq~(\ref{equ:Eq_pixelSpatialAttention}) and reshape it to $W_{map}(i)$\;
    Update $g_{i+1}$ by accumulating the velocity vector in the gradient dirction as:
        \begin{equation}
        g_{i+1}=\mu\cdot{g_{i}}+\frac{\nabla{J}_{x}({x_{i}^{*}, y)}}{||\nabla{J}_{x}({x_{i}^{*}, y)}||_{1}}\times{W_{map}(i)}
        \label{equ:Eq_updateG}
        \end{equation}

    Calculate $\rho_{i}=g_{i+1}\times{\alpha}$\;
    Update $x_{i+1}^{*}=x_{i}^{*}+\rho_{i}$\;
    \If {$||x_{i+1}^{*}-x||_{\infty}>\epsilon$ or $f(x_{i+1}^{*})\neq{y}$}{
        break\;
    }
}
\Return $x^{*}=x_{i+1}^{*}$.
\end{algorithm}

\section{Experiments\label{expriments}}

\textbf{Datasets}: ImageNet\footnote{ImageNet can be download at \emph{http://www.image-net.org/}}.
\\
\textbf{DNN models}: Resnet v2, Inception-v3 (Inc-v3), and Inception Resnet v2(IncRes-v2)~\cite{szegedy2017inception}.
\\
\textbf{Baseline attack methods}: {\small{FGSM~\cite{goodfellow2014explaining}, PGD~\cite{madry2017towards}, C\&W~\cite{carlini2017towards}, MI-FGSM~\cite{dongboosting}, DeepFool~\cite{moosavi2016deepfool}.}} All of the above baselines are implemented by calling foolbox\footnote{ImageNet can be download at \emph{https://foolbox.readthedocs.io/ en/latest/index.html}}.
\\
\textbf{Defense method}: Add Gaussian Blur as a filtering module for deep model, named `defense 1'. Add image transform as a filtering module~\cite{guo2017countering} for deep model, named `defense 2'.
\\
\textbf{Platform}: {\small{i7-7700K 4.20GHzx8 (CPU), TITAN Xp 12GiBx2 (GPU), 16GBx4 memory (DDR4), Ubuntu 16.04 (OS), Python 3.5, Tensorflow-gpu-1.3, Tflearn-0.3.2.

\subsection{Metrics}
In the experiments, mean, 0-norm, 2-norm and infinite-norm of are applied to evaluate the size of the perturbation, which are expressed as $\overline{\rho}$, $L_{0}(\rho)$, $L_{2}(\rho)$ and $L_{\infty}(\rho)$, respectively. And $L_{0}(\cdot)$ distance measurement indicates the number of pixels changed, and the $L_{2}(\cdot)$ represents the Euclidean distance between the two examples, and the $L_{\infty}(\cdot)$ represents the maximum perturbations. And attack success rate (ASR) is used to measure attack ability of adversarial examples.

\subsection{White-box Attack}
Table~\ref{my-label:White_box_Attack} gives the comparison of white-box attack results about untargeted attack, where 5000 images are used for untargeted attack. It is obvious that adversarial examples generated by FineFool have stronger attack ability and smaller perturbation than the baselines. And it can be observed that the perturbations of FineFool is the smallest among all the attack methods under the premise of reaching similar or higher attack effects. As the structure complexity of target deep model increases, the perturbation shows an upward trend, and the perturbation increment of FineFool is also the smallest.

More detailed analysis of Table~\ref{my-label:White_box_Attack}. For untargeted attack against Inc-v3, the ASR (99.9\%) of FineFool is only a little lower than C\&W when other metrics are optimal. And untargeted attack against IncRes-v2, the $L_{2}(\rho)=36.9$ of FineFool is only a little higher than C\&W when other metrics are optimal.

\begin{table}[h]
\centering
\caption{The comparison of untargeted attack results against different DNN models on ImageNet (white-box attack).}
\label{my-label:White_box_Attack}
\resizebox{\linewidth}{!}{
\begin{tabular}{ccccccc}
\hline
\multirow{2}{*}{\textbf{\begin{tabular}[c]{@{}c@{}}Target\\ Model\end{tabular}}} & \multirow{2}{*}{\textbf{Attack}} & \multicolumn{5}{c}{\textbf{Metrics}}                                                                   \\ \cline{3-7}
                                                                                 &                                  & $\overline{\rho}$         & $L_{0}(\rho)$           & $L_{2}(\rho)$         & $L_{\infty}(\rho)$        & ASR                                  \\ \hline
\multirow{5}{*}{Resnet-v2}                                                       & FGSM                             & 0.258          & 81.27\%         & 81.2          & 13.4         & 59.80\%                              \\
                                                                                 & PGD                              & 0.132          & 9.65\%          & 52.2          & 3.5          & 95.40\%                              \\
                                                                                 & C\&W                             & 0.081          & 1.55\%          & 48.9          & 2.7          & 97.00\%                              \\
                                                                                 & MI-FGSM                          & 0.113          & 3.28\%          & 50.2          & 3.2          & 91.20\%                              \\
                                                                                 & \textbf{FineFool}                & \textbf{0.072} & \textbf{1.32\%} & \textbf{46.7} & \textbf{2.6} & \textbf{97.30\%}                     \\ \hline
\multirow{5}{*}{Inc-v3}                                                          & FGSM                             & 0.139          & 86.33\%         & 70.3          & 13.4         & 72.30\%                              \\
                                                                                 & PGD                              & 0.014          & 3.54\%          & 36.9          & 3.5          & 96.00\%                              \\
                                                                                 & C\&W                             & 0.013          & 0.62\%          & 33.0            & 2.7          & \textbf{100.00\%}                    \\
                                                                                 & MI-FGSM                          & 0.013          & 1.94\%          & 34.3          & 3.2          & 99.90\%                              \\
                                                                                 & \textbf{FineFool}                & \textbf{0.002} & \textbf{0.55\%} & \textbf{32.1} & \textbf{2.6} & 99.90\%                              \\ \hline
\multirow{5}{*}{IncRes-v2}                                                       & FGSM                             & 0.145          & 85.21\%         & 80.2          & 19.2         & 61.10\%                              \\
                                                                                 & PGD                              & 0.014          & 3.12\%          & 45.3          & 3.5          & 95.70\%                              \\
                                                                                 & C\&W                             & 0.013          & 1.09\%          & \textbf{36.8} & 3.0            & 99.70\%                              \\
                                                                                 & MI-FGSM                          & 0.021          & 2.12\%          & 36.9          & 3.3          & 98.70\%                              \\
                                                                                 & \textbf{FineFool}                & \textbf{0.012} & \textbf{0.97\%} & 36.9          & \textbf{2.7} & \multicolumn{1}{l}{\textbf{99.80\%}} \\ \hline
\end{tabular}
}
\end{table}

\subsection{Black-box Attack}
It is known that black-box attack is more difficult than white-box. Substitute model is the most common black-box attack method. Table~\ref{my-label:Black_box_Attack} presents the black-box untargeted attack results on ImageNet. Compared with other adversarial attacks, all perturbation metrics of FineFool are the best, indicating that FineFool behaves quite well in complex datasets and target models. The models with different structural complexity are used as substitutes for black-box attack. It can be observed that the closer the substitute model to the target model, the better the attack effect. And black-box attack success rate can be improved when the substitute model is more complex than the target model. Moreover, under the premise that perturbation is as small as possible, the attack effect of FineFool is better than the baselines, which demonstrates its good transfer ability.

More comprehensive description of Table~\ref{my-label:Black_box_Attack}. For untargeted attack against black-box IncRes-v2 based on white-box Resnet-v2, the ASR (50.90\%) of FineFool is lower than ASR (51.30\%) of C\&W. However, improving the ASR of FineFool to 51.30\% via increasing perturbations, the perturbation metrics of FineFool are still optimal.

\begin{table}[h]
\centering
\caption{ The comparison of black-box untargeted attack performance based on substitute models (ImageNet, ¡®*¡¯ denotes white-box
attack).}
\label{my-label:Black_box_Attack}
\resizebox{\linewidth}{!}{
\begin{tabular}{ccccccc}
\hline
\multirow{2}{*}{\textbf{\begin{tabular}[c]{@{}c@{}}Substitute\\ Model\end{tabular}}} & \multirow{2}{*}{\textbf{Attack}} & \multirow{2}{*}{\textbf{$\overline{\rho}$*}} & \multirow{2}{*}{\textbf{$L_{0}(\rho)$*}}  & \multicolumn{3}{c}{\textbf{ASR (black-box)}}                       \\ \cline{5-7}
                                                                                 &                                  &                                    &                                  & Resnet-v2        & Inc-v3           & IncRes-v2        \\ \hline
\multirow{6}{*}{Resnet-v2}                                                       & FGSM                             & 0.258                              & 81.27\%                          & 59.8\%*          & 35.00\%          & 28.20\%          \\
                                                                                 & PGD                              & 0.132                              & 9.65\%                           & 95.4\%*          & 41.20\%          & 37.80\%          \\
                                                                                 & C\&W                             & 0.081                              & 1.55\%                           & 97.0\%*            & 53.20\%          & \textbf{51.30\%} \\
                                                                                 & MI-FGSM                          & 0.113                              & 3.28\%                            & 91.2\%*          & 53.60\%          & 48.80\%          \\
                                                                                 & DeepFool                         & 0.073                            & 1.19\%                           & 95.0\%*            & 42.50\%          & 39.70\%          \\
                                                                                 & \textbf{FineFool}                & \textbf{0.072}                     & \textbf{1.32\%}                  & \textbf{97.3\%*} & \textbf{54.70\%} & 50.90\%          \\ \hline
\multirow{6}{*}{Inc-v3}                                                          & FGSM                             & 0.139                              & 86.33\%                          & 26.70\%          & 72.3\%*          & 28.20\%          \\
                                                                                 & PGD                              & 0.014                              & 3.54\%                           & 38.10\%          & 96.0\%*            & 43.50\%          \\
                                                                                 & C\&W                             & 0.013                              & 0.62\%                           & 45.30\%          & \textbf{100\%*}  & 52.10\%          \\
                                                                                 & MI-FGSM                          & 0.013                              & 1.94\%                           & 39.80\%          & 99.9\%*          & 48.90\%          \\
                                                                                 & DeepFool                         & 0.012                             & 1.28\%                           & 35.70\%          & 98.7\%*          & 36.90\%          \\
                                                                                 & \textbf{FineFool}                & \textbf{0.002}                     & \textbf{0.55\%}                  & \textbf{46.80\%} & 99.9\%*          & \textbf{53.20\%} \\ \hline
\multirow{6}{*}{IncRes-v2}                                                       & FGSM                             & 0.145                              & 85.21\%                          & 27.20\%          & 32.70\%          & 61.1\%*          \\
                                                                                 & PGD                              & 0.014                              & 3.12\%                           & 45.20\%          & 64.90\%          & 95.7\%*          \\
                                                                                 & C\&W                             & 0.013                              & 1.09\%                           & 58.10\%          & 72.10\%          & 99.7\%*          \\
                                                                                 & MI-FGSM                          & 0.021                              & 2.12\%                           & 47.10\%          & 65.60\%          & 98.7\%*          \\
                                                                                 & DeepFool                         & 0.014                              & 12.91\%                          & 42.50\%          & 61.10\%          & 87.5\%*          \\
                                                                                 & \textbf{FineFool}                & \textbf{0.012}                     & \textbf{0.97\%}                  & \textbf{59.40\%} & \textbf{75.40\%} & \textbf{99.8\%*} \\ \hline
\end{tabular}
}
\end{table}

\subsection{Confidence Analysis}
Figures~\ref{fig:perturbation_and_adversary_against_Resnet-v2} and \ref{fig:perturbation_and_adversary_against_Inc-v3} show the visualization of perturbations and adversarial images, which are obtained by different attack methods against different deep models. It is obvious that the perturbations of FineFool is the slightest. And Figure~\ref{fig:confidence_crocodile_Resnet-v2} and \ref{fig:confidence_crocodile_Inc-v3} show the logits curves of different attacks in the iterative process corresponding to Figures~\ref{fig:perturbation_and_adversary_against_Resnet-v2} and \ref{fig:perturbation_and_adversary_against_Inc-v3}, respectively. The solid line with the $\cdot$ denotes logits curve change corresponding to the label of adversarial example, and the dashed line represents logits curve change corresponding to the ground-truth label of original sample. Compared with baselines, it can be seen that the final logits value of adversarial image obtained by FineFool is the highest, the final logits value of original image is the lowest, and the logits curve is linear and convergent.

\begin{figure}[h]
\centering
        \includegraphics[width=0.23\linewidth]{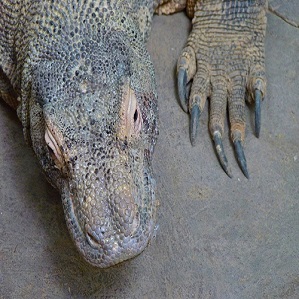}
        \includegraphics[width=0.23\linewidth]{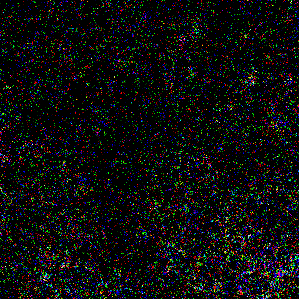}
        \includegraphics[width=0.23\linewidth]{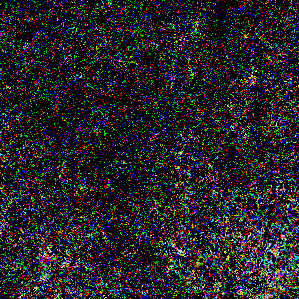}
        \includegraphics[width=0.23\linewidth]{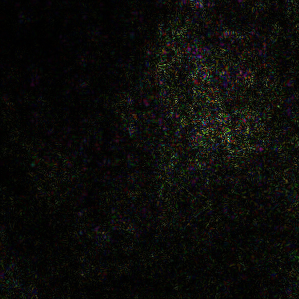} \\
        {~~original~~~~~~~~$\rho_{\rm{MI-FGSM}}$~~~~~~~~~$\rho_{\rm{PGD}}$~~~~~~~~~~~$\rho_{\rm{FineFool}}$}\\
        {~~~~~~~~~~~~~~~~~~~~~~}
        \includegraphics[width=0.23\linewidth]{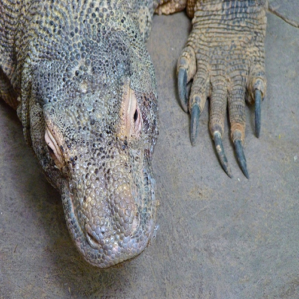}
        \includegraphics[width=0.23\linewidth]{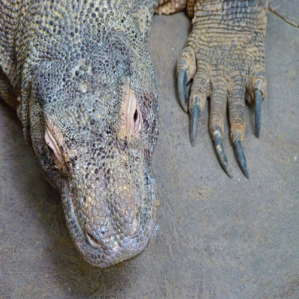}
        \includegraphics[width=0.23\linewidth]{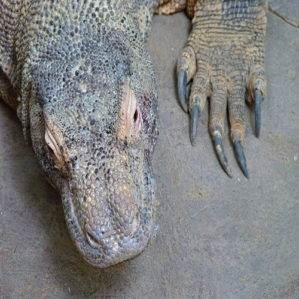} \\
        {~~~~~~~~~~~~~~~~~~~~~~~$Adv_{\rm{MI-FGSM}}$~~~~~~~$Adv_{\rm{PGD}}$~~~~~~$Adv_{\rm{FineFool}}$} \\
    \vspace{-0.2cm}
\caption{The example of different attack methods against Resnet-v2. And all perturbations are magnified 255 times for better visualization.}
\label{fig:perturbation_and_adversary_against_Resnet-v2}
\end{figure}

\begin{figure}[h]
\centering
\includegraphics[width=0.9\linewidth]{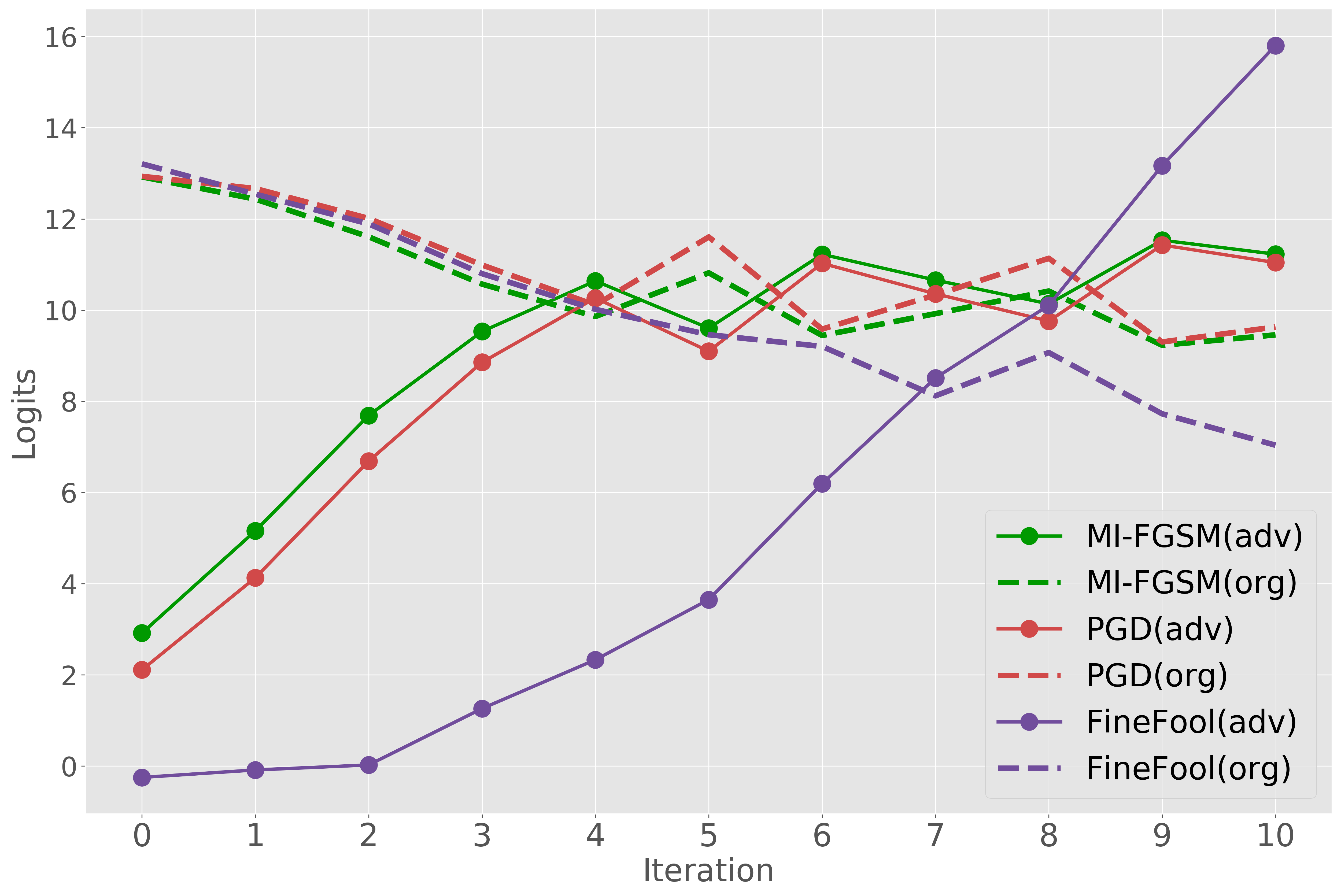}
\caption{The logits curve of different attack methods against Resnet-v2 in iterations.}
\label{fig:confidence_crocodile_Resnet-v2}
\end{figure}

\begin{figure}[h]
\centering
        \includegraphics[width=0.23\linewidth]{orgImage.png}
        \includegraphics[width=0.23\linewidth]{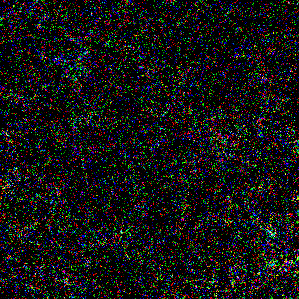}
        \includegraphics[width=0.23\linewidth]{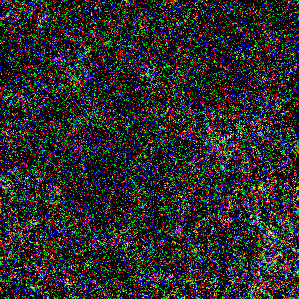}
        \includegraphics[width=0.23\linewidth]{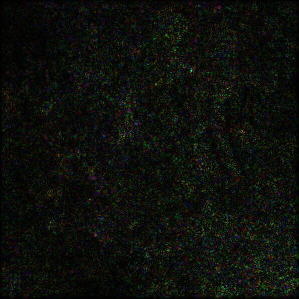} \\
        {~~original~~~~~~~~$\rho_{\rm{MI-FGSM}}$~~~~~~~~~$\rho_{\rm{PGD}}$~~~~~~~~~~~$\rho_{\rm{FineFool}}$}\\
        {~~~~~~~~~~~~~~~~~~~~~~}
        \includegraphics[width=0.23\linewidth]{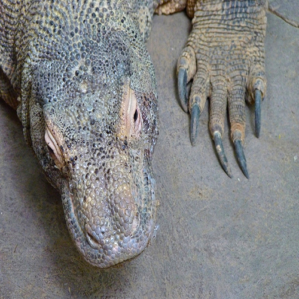}
        \includegraphics[width=0.23\linewidth]{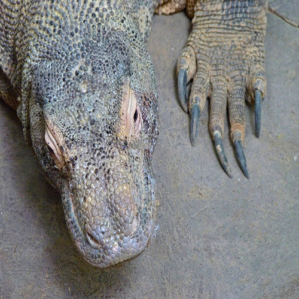}
        \includegraphics[width=0.23\linewidth]{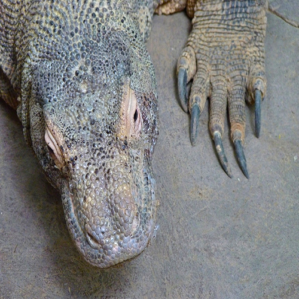} \\
        {~~~~~~~~~~~~~~~~~~~~~~~$Adv_{\rm{MI-FGSM}}$~~~~~~~$Adv_{\rm{PGD}}$~~~~~~$Adv_{\rm{FineFool}}$} \\
    \vspace{-0.2cm}
\caption{The example of different attack methods against Inc-v3. And all perturbations are magnified 255 times for better visualization.}
\label{fig:perturbation_and_adversary_against_Inc-v3}
\end{figure}

\begin{figure}[h]
\centering
\includegraphics[width=0.9\linewidth]{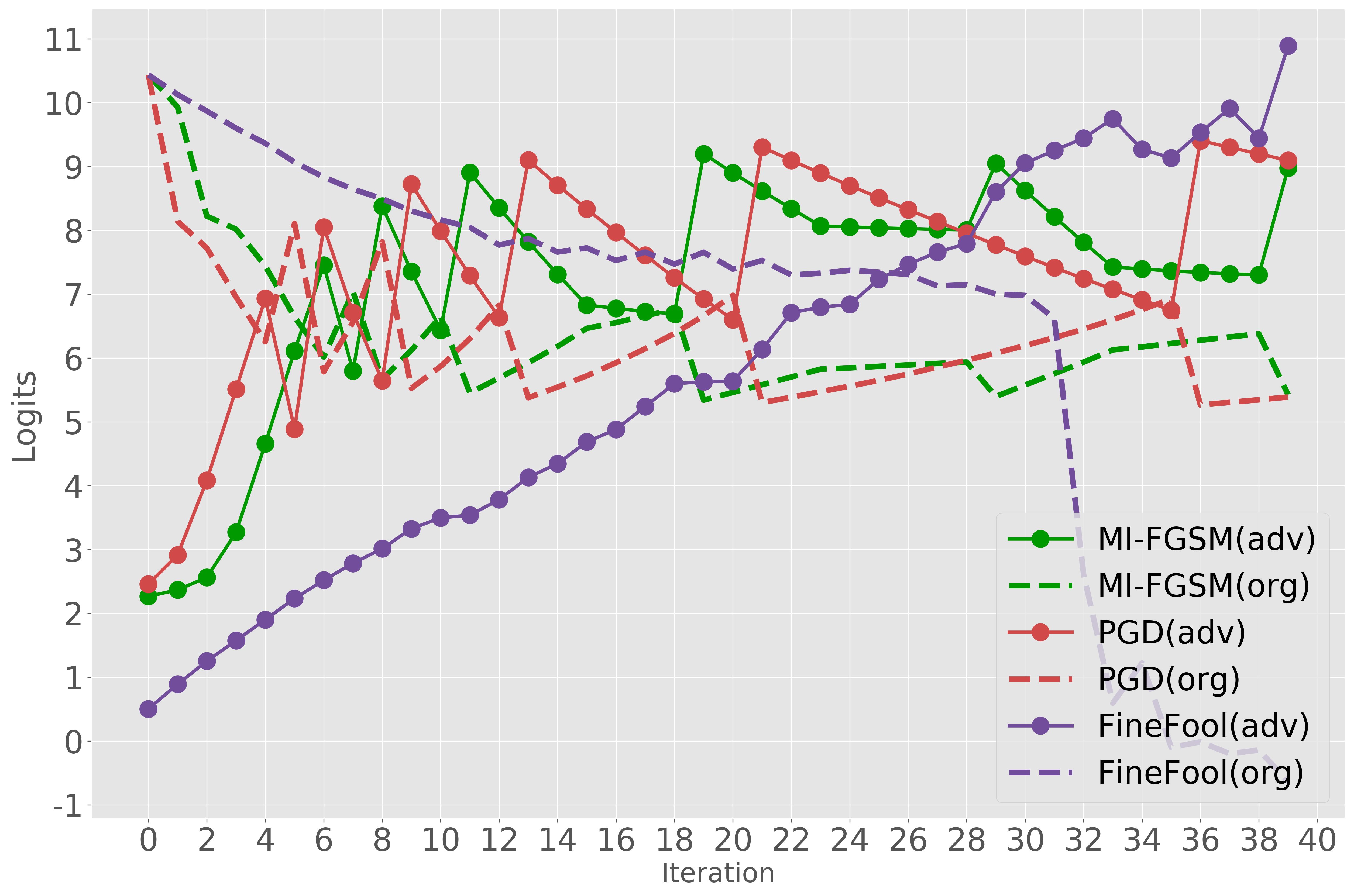}
\caption{The logits curve of different attack methods against Inc-v3 in iterations.}
\label{fig:confidence_crocodile_Inc-v3}
\end{figure}

\subsection{Defense Methods Attack}
Table \ref{my-label:Defense_Methods_Attack} shows the white-box untargeted attack results of different attack methods against Inc-v3 with different defenses on ImageNet. The higher the ASR value, the lower the accuracy and the better the performance of attack method. The Gaussian Blur is applied as one of the defense methods, which's parameter $\sigma$ is 1 and the kernel size is 3*3. Accuracy in Table \ref{my-label:Defense_Methods_Attack} represents the classification accuracy of deep model for adversarial examples generated by attack methods. After the defense strategies are added on deep models, the attack ability of all attack methods is greatly reduced. Gaussian blur can effectively reduce perturbations at the expense of image detail, which is fatal for gradient-based adversarial attacks. At the same time, blur results in the  loss of the contour feature of the target object, which destroys the distribution of attention map. The reconstruction of image transform still retains the basic contour features of the object, so FineFool still keeps stronger attack ability.

\begin{table}[h]
\centering
\caption{The white-box untargeted attack results of different attack methods against Inc-v3 with different defenses on
ImageNet.}
\label{my-label:Defense_Methods_Attack}
\resizebox{\linewidth}{!}{
\begin{tabular}{ccccccc}
\hline
\multirow{2}{*}{\textbf{Attack}} & \multicolumn{2}{c}{\textbf{defense 1}} & \multicolumn{2}{c}{\textbf{defense 2}} & \multicolumn{2}{c}{\textbf{without defense}} \\ \cline{2-7}
                                 & ASR                & accuracy          & ASR                & accuracy          & ASR                    & accuracy            \\ \hline
MI-FGSM                          & 54.40\%            & 65.32\%           & 28.97\%            & 71.52\%           & 99.90\%                & 0.1\%               \\
C\&W                             & 52.30\%            & \textbf{60.87\%}  & 32.89\%            & 70.50\%           & 96.00\%                & 3.9\%               \\
\textbf{FineFool}                & \textbf{55.30\%}   & 65.34\%           & \textbf{58.60\%}   & \textbf{50.86\%}  & \textbf{99.90\%}       & \textbf{0.1\%}      \\ \hline
\end{tabular}
}
\end{table}

\section{Conclusions\label{Conclusions}}
In this paper, we try to put forward a novel adversarial attack method based on attention mechanism, FineFool, which pay more attention to critical classification features and object contour. In the course of adversarial attack research, an important observation has been made by visualization that there are strong correlations between perturbations and object contour. In order to produce more effective adversarial examples, slighter perturbations are implemented by focusing features onto object contour. The key to FineFool is channel spatial attention and pixel spatial attention, which focuses on channel feature distribution and pixel region distribution, respectively. Compared with the current state-of-the-art attacks, the perturbations of FineFool can be further reduced while maintaining similar attack success rate. From theoretical explanation and experimental analysis, we can conclude that FineFool is capable of generation of high effective adversarial examples with both better attack performance and less perturbations. Intensive experiments on ImageNet dataset show that the proposed method is attack effective against different deep models with defensive network add-on, which also can be applied to evaluate robustness of deep models and defense methods.


\small
\bibliographystyle{aaai}
\bibliography{myref}

\begin{thebibliography}{}

\bibitem[\protect\citeauthoryear{Akhtar and Mian}{2018}]{akhtar2018threat}
Akhtar, N., and Mian, A.
\newblock 2018.
\newblock Threat of adversarial attacks on deep learning in computer vision: A
  survey.
\newblock {\em arXiv preprint arXiv:1801.00553}.

\bibitem[\protect\citeauthoryear{Akhtar, Liu, and
  Mian}{2017}]{akhtar2017defense}
Akhtar, N.; Liu, J.; and Mian, A.
\newblock 2017.
\newblock Defense against universal adversarial perturbations.
\newblock {\em arXiv preprint arXiv:1711.05929}.

\bibitem[\protect\citeauthoryear{Bhagoji \bgroup et al\mbox.\egroup
  }{2017}]{Arjun2017Enhancing}
Bhagoji, A.~N.; Cullina, D.; Sitawarin, B.; and Mittal, P.
\newblock 2017.
\newblock Enhancing robustness of machine learning systems via data
  transformations.

\bibitem[\protect\citeauthoryear{Carlini and Wagner}{2017}]{carlini2017towards}
Carlini, N., and Wagner, D.
\newblock 2017.
\newblock Towards evaluating the robustness of neural networks.
\newblock In {\em Security and Privacy (SP), 2017 IEEE Symposium on},  39--57.
\newblock IEEE.

\bibitem[\protect\citeauthoryear{Chen \bgroup et al\mbox.\egroup
  }{2017}]{chen2017sca}
Chen, L.; Zhang, H.; Xiao, J.; Nie, L.; Shao, J.; Liu, W.; and Chua, T.-S.
\newblock 2017.
\newblock Sca-cnn: Spatial and channel-wise attention in convolutional networks
  for image captioning.
\newblock In {\em Computer Vision and Pattern Recognition (CVPR), 2017 IEEE
  Conference on},  6298--6306.
\newblock IEEE.

\bibitem[\protect\citeauthoryear{Cubuk \bgroup et al\mbox.\egroup
  }{2017}]{cubuk2017intriguing}
Cubuk, E.~D.; Zoph, B.; Schoenholz, S.~S.; and Le, Q.~V.
\newblock 2017.
\newblock Intriguing properties of adversarial examples.
\newblock {\em arXiv preprint arXiv:1711.02846}.

\bibitem[\protect\citeauthoryear{Das \bgroup et al\mbox.\egroup
  }{2017}]{das2017keeping}
Das, N.; Shanbhogue, M.; Chen, S.-T.; Hohman, F.; Chen, L.; Kounavis, M.~E.;
  and Chau, D.~H.
\newblock 2017.
\newblock Keeping the bad guys out: Protecting and vaccinating deep learning
  with jpeg compression.
\newblock {\em arXiv preprint arXiv:1705.02900}.

\bibitem[\protect\citeauthoryear{Dong \bgroup et al\mbox.\egroup
  }{2017a}]{dongboosting}
Dong, Y.; Liao, F.; Pang, T.; Su, H.; Hu, X.; Li, J.; and Zhu, J.
\newblock 2017a.
\newblock Boosting adversarial attacks with momentum.

\bibitem[\protect\citeauthoryear{Dong \bgroup et al\mbox.\egroup
  }{2017b}]{dong2017towards}
Dong, Y.; Su, H.; Zhu, J.; and Bao, F.
\newblock 2017b.
\newblock Towards interpretable deep neural networks by leveraging adversarial
  examples.
\newblock {\em arXiv preprint arXiv:1708.05493}.

\bibitem[\protect\citeauthoryear{Evtimov \bgroup et al\mbox.\egroup
  }{2017}]{evtimov2017robust}
Evtimov, I.; Eykholt, K.; Fernandes, E.; Kohno, T.; Li, B.; Prakash, A.;
  Rahmati, A.; and Song, D.
\newblock 2017.
\newblock Robust physical-world attacks on deep learning models.
\newblock {\em arXiv preprint arXiv:1707.08945} 1.

\bibitem[\protect\citeauthoryear{Gao \bgroup et al\mbox.\egroup
  }{2017}]{gao2017deepcloak}
Gao, J.; Wang, B.; Lin, Z.; Xu, W.; and Qi, Y.
\newblock 2017.
\newblock Deepcloak: Masking deep neural network models for robustness against
  adversarial samples.

\bibitem[\protect\citeauthoryear{Goodfellow, Shlens, and
  Szegedy}{2014}]{goodfellow2014explaining}
Goodfellow, I.~J.; Shlens, J.; and Szegedy, C.
\newblock 2014.
\newblock Explaining and harnessing adversarial examples.
\newblock {\em arXiv preprint arXiv:1412.6572}.

\bibitem[\protect\citeauthoryear{Gu and Rigazio}{2014}]{gu2014towards}
Gu, S., and Rigazio, L.
\newblock 2014.
\newblock Towards deep neural network architectures robust to adversarial
  examples.
\newblock {\em arXiv preprint arXiv:1412.5068}.

\bibitem[\protect\citeauthoryear{Guo \bgroup et al\mbox.\egroup
  }{2017}]{guo2017countering}
Guo, C.; Rana, M.; Ciss{\'e}, M.; and van~der Maaten, L.
\newblock 2017.
\newblock Countering adversarial images using input transformations.
\newblock {\em arXiv preprint arXiv:1711.00117}.

\bibitem[\protect\citeauthoryear{Jacob \bgroup et al\mbox.\egroup
  }{2018}]{Buckman2018Thermometer}
Jacob, B.; Aurko, R.; Colin, R.; and Ian, G.
\newblock 2018.
\newblock Thermometer encoding: One hot way to resist adversarial examples.
\newblock  1--22.

\bibitem[\protect\citeauthoryear{Koh and Liang}{2017}]{koh2017understanding}
Koh, P.~W., and Liang, P.
\newblock 2017.
\newblock Understanding black-box predictions via influence functions.
\newblock {\em arXiv preprint arXiv:1703.04730}.

\bibitem[\protect\citeauthoryear{Kurakin, Goodfellow, and
  Bengio}{2016}]{kurakin2016adversarialPhysical}
Kurakin, A.; Goodfellow, I.; and Bengio, S.
\newblock 2016.
\newblock Adversarial examples in the physical world.
\newblock {\em arXiv preprint arXiv:1607.02533}.

\bibitem[\protect\citeauthoryear{Lee, Han, and Lee}{2017}]{lee2017generative}
Lee, H.; Han, S.; and Lee, J.
\newblock 2017.
\newblock Generative adversarial trainer: Defense to adversarial perturbations
  with gan.
\newblock {\em arXiv preprint arXiv:1705.03387}.

\bibitem[\protect\citeauthoryear{Madry \bgroup et al\mbox.\egroup
  }{2017}]{madry2017towards}
Madry, A.; Makelov, A.; Schmidt, L.; Tsipras, D.; and Vladu, A.
\newblock 2017.
\newblock Towards deep learning models resistant to adversarial attacks.
\newblock {\em arXiv preprint arXiv:1706.06083}.

\bibitem[\protect\citeauthoryear{Meng and Chen}{2017}]{meng2017magnet}
Meng, D., and Chen, H.
\newblock 2017.
\newblock Magnet: a two-pronged defense against adversarial examples.
\newblock In {\em Proceedings of the 2017 ACM SIGSAC Conference on Computer and
  Communications Security},  135--147.
\newblock ACM.

\bibitem[\protect\citeauthoryear{Miyato \bgroup et al\mbox.\egroup
  }{2017}]{miyato2017virtual}
Miyato, T.; Maeda, S.-i.; Koyama, M.; and Ishii, S.
\newblock 2017.
\newblock Virtual adversarial training: a regularization method for supervised
  and semi-supervised learning.
\newblock {\em arXiv preprint arXiv:1704.03976}.

\bibitem[\protect\citeauthoryear{Miyato, Dai, and
  Goodfellow}{2016}]{miyato2016adversarial}
Miyato, T.; Dai, A.~M.; and Goodfellow, I.
\newblock 2016.
\newblock Adversarial training methods for semi-supervised text classification.
\newblock {\em arXiv preprint arXiv:1605.07725}.

\bibitem[\protect\citeauthoryear{Moosavi~Dezfooli, Fawzi, and
  Frossard}{2016}]{moosavi2016deepfool}
Moosavi~Dezfooli, S.~M.; Fawzi, A.; and Frossard, P.
\newblock 2016.
\newblock Deepfool: a simple and accurate method to fool deep neural networks.
\newblock In {\em Proceedings of 2016 IEEE Conference on Computer Vision and
  Pattern Recognition (CVPR)}, number EPFL-CONF-218057.

\bibitem[\protect\citeauthoryear{Nayebi and
  Ganguli}{2017}]{nayebi2017biologically}
Nayebi, A., and Ganguli, S.
\newblock 2017.
\newblock Biologically inspired protection of deep networks from adversarial
  attacks.
\newblock {\em arXiv preprint arXiv:1703.09202}.

\bibitem[\protect\citeauthoryear{Papernot \bgroup et al\mbox.\egroup
  }{2016}]{papernot2016distillation}
Papernot, N.; McDaniel, P.; Wu, X.; Jha, S.; and Swami, A.
\newblock 2016.
\newblock Distillation as a defense to adversarial perturbations against deep
  neural networks.
\newblock In {\em Security and Privacy (SP), 2016 IEEE Symposium on},
  582--597.
\newblock IEEE.

\bibitem[\protect\citeauthoryear{Ross and
  Doshi-Velez}{2017}]{ross2017improving}
Ross, A.~S., and Doshi-Velez, F.
\newblock 2017.
\newblock Improving the adversarial robustness and interpretability of deep
  neural networks by regularizing their input gradients.
\newblock {\em arXiv preprint arXiv:1711.09404}.

\bibitem[\protect\citeauthoryear{Rozsa, Gunther, and
  Boult}{2016}]{rozsa2016towards}
Rozsa, A.; Gunther, M.; and Boult, T.~E.
\newblock 2016.
\newblock Towards robust deep neural networks with bang.
\newblock {\em arXiv preprint arXiv:1612.00138}.

\bibitem[\protect\citeauthoryear{Sun, Ozay, and
  Okatani}{2017}]{sun2017hypernetworks}
Sun, Z.; Ozay, M.; and Okatani, T.
\newblock 2017.
\newblock Hypernetworks with statistical filtering for defending adversarial
  examples.
\newblock {\em arXiv preprint arXiv:1711.01791}.

\bibitem[\protect\citeauthoryear{Szegedy \bgroup et al\mbox.\egroup
  }{2013}]{szegedy2013intriguing}
Szegedy, C.; Zaremba, W.; Sutskever, I.; Bruna, J.; Erhan, D.; Goodfellow, I.;
  and Fergus, R.
\newblock 2013.
\newblock Intriguing properties of neural networks.
\newblock {\em arXiv preprint arXiv:1312.6199}.

\bibitem[\protect\citeauthoryear{Szegedy \bgroup et al\mbox.\egroup
  }{2017}]{szegedy2017inception}
Szegedy, C.; Ioffe, S.; Vanhoucke, V.; and Alemi, A.~A.
\newblock 2017.
\newblock Inception-v4, inception-resnet and the impact of residual connections
  on learning.
\newblock In {\em AAAI}, volume~4, ~12.

\bibitem[\protect\citeauthoryear{Tanay and Griffin}{2016}]{tanay2016boundary}
Tanay, T., and Griffin, L.
\newblock 2016.
\newblock A boundary tilting persepective on the phenomenon of adversarial
  examples.
\newblock {\em arXiv preprint arXiv:1608.07690}.

\bibitem[\protect\citeauthoryear{Wan \bgroup et al\mbox.\egroup
  }{2018}]{Wan2018Rethinking}
Wan, W.; Zhong, Y.; Li, T.; and Chen, J.
\newblock 2018.
\newblock Rethinking feature distribution for loss functions in image
  classification.
\newblock  1--10.

\bibitem[\protect\citeauthoryear{Wang \bgroup et al\mbox.\egroup
  }{2017}]{wang2017residual}
Wang, F.; Jiang, M.; Qian, C.; Yang, S.; Li, C.; Zhang, H.; Wang, X.; and Tang,
  X.
\newblock 2017.
\newblock Residual attention network for image classification.
\newblock {\em arXiv preprint arXiv:1704.06904}.

\bibitem[\protect\citeauthoryear{Xie \bgroup et al\mbox.\egroup
  }{2017}]{xie2017adversarial}
Xie, C.; Wang, J.; Zhang, Z.; Zhou, Y.; Xie, L.; and Yuille, A.
\newblock 2017.
\newblock Adversarial examples for semantic segmentation and object detection.
\newblock In {\em International Conference on Computer Vision. IEEE}.

\bibitem[\protect\citeauthoryear{Xu \bgroup et al\mbox.\egroup
  }{2015}]{xu2015show}
Xu, K.; Ba, J.; Kiros, R.; Cho, K.; Courville, A.; Salakhudinov, R.; Zemel, R.;
  and Bengio, Y.
\newblock 2015.
\newblock Show, attend and tell: Neural image caption generation with visual
  attention.
\newblock In {\em International conference on machine learning},  2048--2057.

\bibitem[\protect\citeauthoryear{Yang \bgroup et al\mbox.\egroup
  }{2016}]{yang2016stacked}
Yang, Z.; He, X.; Gao, J.; Deng, L.; and Smola, A.
\newblock 2016.
\newblock Stacked attention networks for image question answering.
\newblock In {\em Proceedings of the IEEE Conference on Computer Vision and
  Pattern Recognition},  21--29.

\bibitem[\protect\citeauthoryear{Zhang \bgroup et al\mbox.\egroup
  }{2018}]{zhang2018self}
Zhang, H.; Goodfellow, I.; Metaxas, D.; and Odena, A.
\newblock 2018.
\newblock Self-attention generative adversarial networks.
\newblock {\em arXiv preprint arXiv:1805.08318}.

\bibitem[\protect\citeauthoryear{Zheng \bgroup et al\mbox.\egroup
  }{2016}]{zheng2016improving}
Zheng, S.; Song, Y.; Leung, T.; and Goodfellow, I.
\newblock 2016.
\newblock Improving the robustness of deep neural networks via stability
  training.
\newblock In {\em Proceedings of the IEEE Conference on Computer Vision and
  Pattern Recognition},  4480--4488.

\end{thebibliography}

\end{document}